\documentclass{article}

\usepackage[nonatbib,preprint]{neurips_2020}
\usepackage[numbers]{natbib}
\usepackage[utf8]{inputenc} 
\usepackage[T1]{fontenc}    
\usepackage{hyperref}       
\usepackage{url}            
\usepackage{booktabs}       
\usepackage{amsfonts, amsmath}       
\usepackage{nicefrac}       
\usepackage{microtype}      
\usepackage{bbm}

\usepackage{amsmath,amsfonts,amssymb,amsthm, color}
\usepackage{algorithm,algorithmicx}
\usepackage{epsfig,wrapfig,epsf,subcaption}
\usepackage[outercaption]{sidecap} 
\usepackage{times}
\usepackage{fancyvrb}
\usepackage{enumerate}
\usepackage{tikz}
\usepackage{bm}
\usetikzlibrary{arrows,shapes}
\usetikzlibrary{patterns}
\usepackage{graphicx}
\usepackage{comment}
\usepackage{ltxtable}
\usepackage{nicefrac}
\usepackage{multirow}
\usepackage{setspace}
\usepackage[noend]{algpseudocode}
\usepackage[utf8]{inputenc} 
\usepackage[T1]{fontenc}    
\usepackage{url}            
\usepackage{booktabs}       
\usepackage{microtype}      
\usepackage{pbox}
\usepackage{xfrac}

\makeatletter
\newtheorem*{rep@theorem}{\rep@title}
\newcommand{\newreptheorem}[2]{%
\newenvironment{rep#1}[1]{%
 \def\rep@title{#2 \ref{##1}}%
 \begin{rep@theorem}}%
 {\end{rep@theorem}}}
\makeatother

\newreptheorem{theorem}{Theorem}
\newreptheorem{lemma}{Lemma}
\newreptheorem{definition}{Definition}

\theoremstyle{definition}

\newcounter{saveenumi}

\usepackage{amsmath,amsfonts,bm}

\def\1{\bm{1}}

\DeclareMathAlphabet{\mathsfit}{\encodingdefault}{\sfdefault}{m}{sl}
\SetMathAlphabet{\mathsfit}{bold}{\encodingdefault}{\sfdefault}{bx}{n}

\usepackage{graphicx}
\usepackage[mathscr]{eucal}
\usepackage{enumitem}

\title{Improving Calibration in Deep Metric Learning With Cross-Example Softmax}

\author{Andreas Veit\thanks{These two authors contributed equally} \quad Kimberly Wilber$^*$
\\ Google Research
\\ \texttt{\{aveit,kwilber\}@google.com}}

\begin{document}

\maketitle

\begin{abstract}
Modern image retrieval systems increasingly rely on the use of deep neural networks to learn embedding spaces in which distance encodes the relevance between a given query and image. In this setting, existing approaches tend to emphasize one of two properties. Triplet-based methods capture \emph{top-$k$ relevancy}, where all top-$k$ scoring documents are assumed to be relevant to a given query Pairwise contrastive models capture \emph{threshold relevancy}, where all documents scoring higher than some threshold are assumed to be relevant. In this paper, we propose \emph{Cross-Example Softmax} which combines the properties of top-$k$ and threshold relevancy. In each iteration, the proposed loss encourages all queries to be closer to their matching images than all queries are to all non-matching images. This leads to a globally more calibrated similarity metric and makes distance more interpretable as an absolute measure of relevance. We further introduce \emph{Cross-Example Negative Mining}, in which each pair is compared to the hardest negative comparisons across the entire batch. Empirically, we show in a series of experiments on Conceptual Captions and Flickr30k, that the proposed method effectively improves global calibration and also retrieval performance.
\end{abstract}

\section{Introduction}
The goal of large-scale information retrieval is to efficiently find relevant documents for a given query among potentially billions of candidates. A canonical example is image search: finding relevant images given a text query. One common implementation is to learn a real-valued scoring function to rank all images. Since using large neural networks to compute the relevance of each query-image pair is prohibitively expensive, recent deep learning systems encode semantic relevance as distance in a vector space. These systems can model complex semantic relationships while still allowing for efficient retrieval using approximate nearest neighbor search through a large database of images. 

\begin{figure}[t]
\centering
\includegraphics[width=0.65\linewidth]{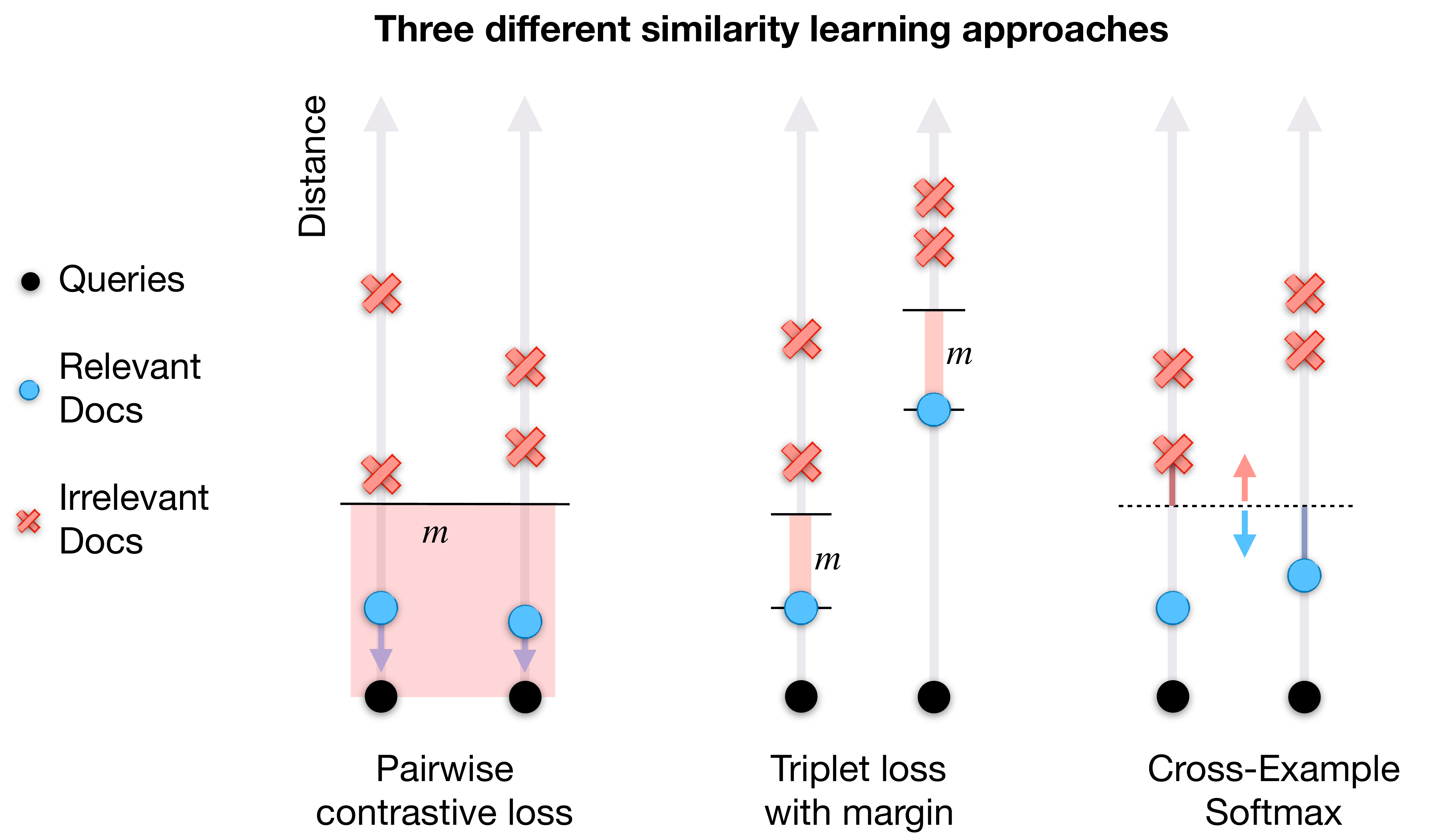}
\caption{
Left: Pairwise contrastive loss minimizes the distance of matching query/document pairs while pushing nonmatching pairs up to some pre-selected margin. 
Center: Triplet loss constrains non-matching documents to be at least some margin further away than matching documents for a given reference query.
Right: The proposed Cross-Example Softmax compares distances across queries, encouraging all matching query/document pairs to be closer than all non-matching pairs.
}
\label{fig:intro}
\vspace{-5pt}
\end{figure}

In this embedding based retrieval setting, two approaches are commonly used to determine the relevance of retrieved documents. In \emph{top-$k$ relevancy}, all top-$k$ scoring documents are assumed to be relevant to a given query, while in \emph{threshold relevancy}, all documents scoring higher than some threshold are assumed to be relevant. Both properties are commonly used; \emph{top-$k$ relevancy} is useful when searching for a document known to exist, \emph{threshold relevancy} is useful when it is not known whether any relevant document exists. Although both properties are desirable, common learning approaches tend to explicitly optimize for only one of these properties. To address this issue, we propose a novel learning approach that simultaneously optimizes both properties.

Early approaches for embedding learning used pairwise \emph{contrastive losses}~\cite{siamese}, where the distance between matching query/document pairs is minimized and the distance between non-matching pairs is maximized until they reach a specified margin. Using explicit definitions for what constitutes matching and non-matching pairs, these approaches have been an effective means to achieve threshold relevancy. However, the binary nature of these pairwise approaches limits their ability to model the relative ordering among the matching pairs, limiting their effectiveness for top-$k$ relevancy.

To address this challenge, relative comparison approaches such as triplet loss~\cite{facenet} have been proposed that explicitly model the relative ordering of pairwise query/document distances with respect to a single reference query. While effectively modeling top-$k$ relevancy, the query-specific conditioning of pairwise distances implies that absolute distances are not necessarily meaningful across different queries, limiting threshold relevancy. Recent approaches commonly go beyond individual triplets with the usage of \emph{Sampled Softmax}~\cite{sampledsoftmax}. There, a batch of matching query/document pairs are embedded into their vector representations. Then, for each query, the documents from the other pairs in the batch can be re-used as negative comparisons. Since most of these random documents are unlikely to be informative to each query, it is also common to use stochastic negative mining~\cite{snm} to only use the most informative negative documents, i.e., with the highest similarity score.

In this paper, we propose \emph{Cross-Example Softmax}, an extension to Sampled Softmax which combines the properties of top-$k$ relevancy and threshold relevancy by introducing cross-example comparisons. Specifically, in addition to ranking a positive query/document pair over all other within-batch pairs containing the same query, Cross-Example Softmax also ranks each positive pair over all non-matching pairs containing different queries. Figure~\ref{fig:intro} illustrates the effect of Cross-Example Softmax and how it leads to calibrated distance scores.
The proposed method further allows us to extend the concept of stochastic negative mining to \emph{Cross-Example Negative Mining}. There, instead of only mining the most informative negative documents for the given query, we select those non-matching pairs with the highest similarity scores across all queries.

In experiments on Conceptual Captions and Flickr30k, we demonstrate that Cross-Example Softmax effectively learns embedding spaces that are more globally calibrated and that have higher retrieval performance. Quantitatively, in terms of area under the precision-recall curve, this results in an improvement from 14.6 to 20.1 on the Conceptual Captions test set. In terms of Recall@1, we observe improvements from 25.87\% to 26.91\% on Conceptual Captions and from 29.22\% to 30.49\% on Flickr30k.

\noindent
Overall, we make the following \textbf{main contributions:}
\begin{itemize}
\item With Cross-Example Softmax, we propose a novel loss function for embedding learning which combines the properties of top-$k$ and threshold relevancy.
\item We further introduce Cross-Example Negative Mining, where the per-example loss is focused on the hardest incorrect query/document pairs across the entire batch.
\item In experiments on Conceptual Captions and Flickr30k, we show that the proposed methods effectively improve global score calibration and retrieval performance.
\end{itemize}

\section{Related Work}
\textbf{Vision-Language Models.} Approaches that combine the visual and language modalities have been of great interest in recent years, addressing tasks such as image captioning~\cite{cococaptions,showandtell,showattendandtell}, visual question answering~\cite{vqa}, visually-grounded dialog~\cite{visualdialog} and more. One common characteristic of many of these approaches is to combine features from both modalities early in the model~\cite{laurensvqa}. Other work trains models that merge visual and language features in Transformer-based~\cite{transformer} architectures~\cite{fusionbert,unicoder,visualbert,vilbert,vlbert,lxmert}. These approaches are powerful because the model can consider nonlinear interactions between vision and language features. However, they are not suited to our large-scale image retrieval task, because inferring a pairwise similarity score between a query and millions of documents becomes intractable. Our work instead focuses on deep metric learning using a factorized model with two separate image and text encoders, limiting the interaction between modalities to an inner product.

\textbf{Deep Metric Learning.} Metric learning using deep models is a well-studied problem with many applications~\cite{extreme-classification,fashion-retrieval,facenet,fashion}, especially where the output space is very large. Early approaches are based upon Siamese networks~\cite{siamese} with contrastive loss on pairwise data or relative triplet similarity comparisons~\cite{tripletnet,facenet}. Inspired by the success of large-scale classification tasks on ImageNet~\cite{imagenet}, more recent models are mainly trained using sampled softmax loss with cross entropy~\cite{sampled-softmax-bengio,sampled-softmax-jean}. Recently, several works have proposed modifications to the sampled softmax loss, by normalization~\cite{angular-softmax}, adding margins~\cite{lsoftmax,additivemargin,cosface}, and tuning the scaling temperature~\cite{adacos,heatedupsoftmax}. Differing from our work, all of these works focus on optimizing the relative ordering of labels given an anchor query. In contrast, the method proposed in our work optimizes for \emph{each} input query the score of the correct label against the \emph{entire distribution of all possible negative query/label pairs in the entire batch}, even across queries.

\textbf{Stochastic Negative Mining.}
In the setting of large output spaces, for any given query, most documents are not relevant and thus including them in the loss function is not informative for the optimization. To address this challenge, several works have proposed to mine for the hardest and most informative negative labels~\cite{w2v,snm}. However, as an approximation of per-query softmax, these methods perform negative mining only with respect to one single query at a time. In our work, we propose negative example mining across examples, wherein we mine for the globally hardest negative query/document comparison in the batch.

\textbf{Score Calibration.}
There has been a sustained interest in score calibration to ensure scores are consistently normalized or interpretable~\cite{platt,weibull,meta-recognition}. One common approach is to interpret the output of the softmax function applied to model logits as probabilities. While the output of a softmax is technically a probability distribution in the sense that it is normalized, computing the probability for any label also requires the comparison to all other labels. In the setting of large output spaces, this is generally not possible, because the probability space is too large to calculate. To address this challenge, in this paper we propose a new loss function that explicitly encourages the underlying logits to be calibrated. This is done during training, not as a post-recognition calibration step as in~\cite{platt,meta-recognition}. This allows the comparison of label scores across queries without needing to compute scores for all other labels.

\section{Method}
Consider the multiclass classification setting with a sample of instances $x \in \mathcal{X}$ and their associated labels $y \in \mathscr{Y}$ with $|\mathscr{Y}| = K$. There, the goal is to learn a scoring function $f : \mathscr{X} \rightarrow \mathbb{R}^K$ that is able to score the labels for each instance according to their relevance. The information retrieval setting at hand can be defined analogous with a set of queries $\mathcal{X}$ and associated relevant documents $\mathscr{Y}$. Our goal is to learn a scoring function which can sort all documents according to their relevance for a given query.

In our text-to-image retrieval setting, $x_i$ is a text query and $y_i$ is its corresponding relevant image. We aim to learn a text encoder $f_{text} : (x_i) \rightarrow \mathbf{x}_{i} \in \mathbb{R}^d$ and an image encoder $f_{imange} : (y_i) \rightarrow \mathbf{y}_{i} \in \mathbb{R}^d$ that project the text and image into a shared $d$-dimensional embedding space. The relevance score between query $x_i$ and image $y_j$ is the dot product between their vector representations $s_{i,j} = \langle \mathbf{x}_{i}, \mathbf{y}_{i} \rangle$.

\begin{figure}[t]
\centering
\includegraphics[width=1\linewidth]{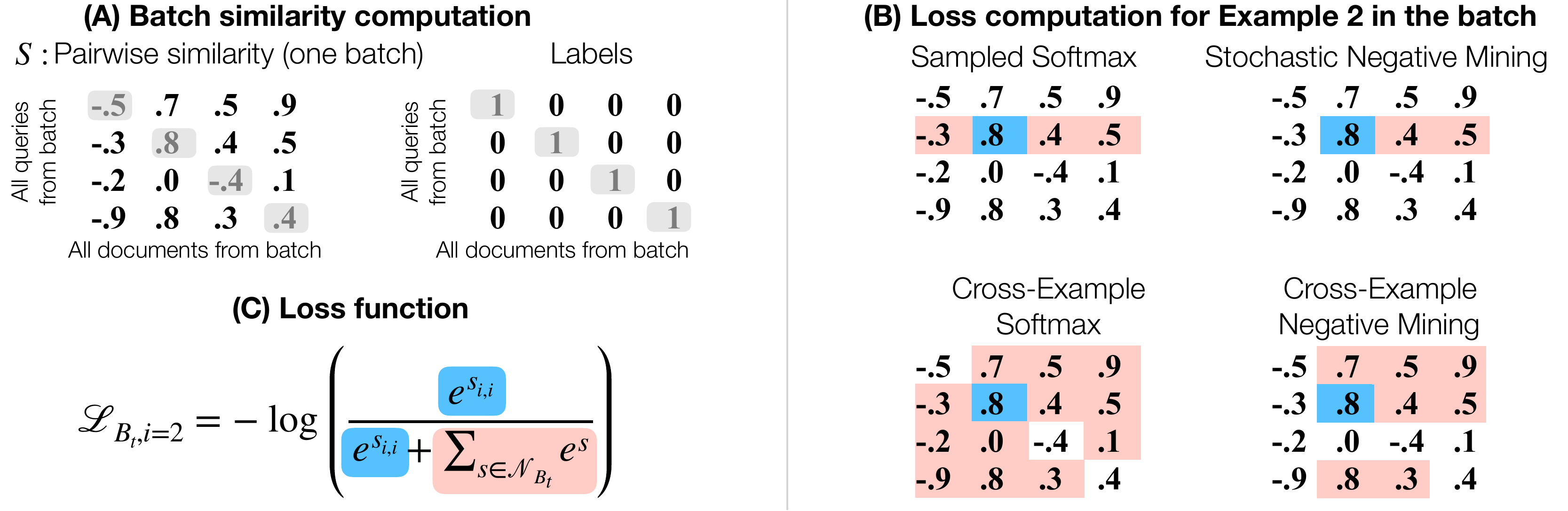}
\caption{\textbf{Cross-Example Softmax} operates on a pairwise distance matrix~(A) computed from a batch of paired queries and documents. Comparisons to other methods are shown in~(B). In Sampled Softmax, for a given query the similarity score for the positive document is compared to all other documents within the batch. Stochastic Negative Mining focuses the loss on the hardest document for each query. For Cross-Example Softmax, each positive query/document pair is compared to all non-matching pairs across the entire batch. Cross-Example Negative Mining focuses the loss on the hardest non-matching query/document pairs across the entire batch. In all of these approaches, negatives appear in the partition function of the softmax activation~(C).
\label{fig:sampling-methods}}
\vspace{-5pt}
\end{figure}

\textbf{Sampled Softmax.}
In the standard multiclass classification setting, a Softmax Cross-Entropy loss over the whole label space is used to optimize the model. Ideally, in the retrieval setting we would also compare the score of a matching document to all other documents in the database. 
However, since the number of documents may be in the billions, the Softmax Cross-Entropy loss is commonly only computed over a random subset of labels (Sampled Softmax).

Specifically, consider a mini-batch of $N$ corresponding query/document pairs $B_t = \{ (x_1, y_1), \ldots, (x_N, y_N) \}$. Given the vector representations of all text queries and images from the mini-batch, we can compute the pairwise similarity matrix between all possible pairs $S \in \mathbb{R}^{N\times N} = s_{i,j}, \forall \ i,j \in [1,...,N]$. An example of such a matrix for $N=4$ is illustrated on the left in Fig.~\ref{fig:sampling-methods}~(A). 
It is commonly assumed that for a given query $x_i$, within the batch only its corresponding document $y_i$ is relevant. All other documents $y_j, j\neq i$ within the same batch are assumed to be irrelevant to that query. The matching relationship between queries and documents within a batch is illustrated on the right in Fig.~\ref{fig:sampling-methods}~(A) with $1$ indicating a matching relationship and $0$ indicating a non-match. Formally, let $\mathcal{N}_{i,B_t}$ be the set of similarity scores between query $x_i$ and all non-matching documents in the batch, i.e., except for the query's corresponding document $y_i$. In Fig.~\ref{fig:sampling-methods} this would be all scores in row $i$ of $S$ except the relevant document.

\begin{equation}
    \mathcal{N}_{i,B_t} = \{ s_{i,j} : j \neq i \}
\end{equation}

With this we can define \emph{Sampled Softmax} Cross-Entropy as a relative ranking loss between the relevance score of a query and its matching document $s_{i,i}$ and its relevance scores to all non-matching documents $s \in \mathcal{N}_{i, B_t}$. Formally,
\begin{equation}
    \mathcal{L}_{B_t} = - \frac{1}{N} \sum_{i=1}^{N}
\log\left(\frac{e^{s_{i,i}}}{e^{s_{i,i}} + \sum_{s\;\in\;\mathcal{N}_{i, B_t}} e^{s}} \right)
\label{eq:SampledSoftmax}
\end{equation}

Fig.~\ref{fig:sampling-methods}~(B) in the top left illustrates the per-example loss for the second query. The score of the matching pair $s_{2,2}$ is emphasized in blue and $\mathcal{N}_{2, B_t}$ in marked in red. The figure highlights that the loss only considers pairs from the same query. 

\textbf{Stochastic Negative Mining.}
For Sampled Softmax a relevant document is only compared to a small subset of random documents. Most of these documents can be easily recognized as irrelevant to the query, limiting their informativeness to the optimization.

As a consequence, Stochastic Negative Mining has been proposed~\cite{snm}, wherein one only selects the most difficult negative documents for each query within the batch. Formally, let $\text{top}k( \mathcal{N}_{i, B_t})$ be the set of the top $k$ largest scores within the set of negative scores for query $x_i$.
With this, we can write the modified loss defined only over the most difficult documents for each query as
\begin{equation}
    \mathcal{L}_{B_t} = - \frac{1}{N} \sum_{i=1}^{N}
\log\left(\frac{e^{s_{i,i}}}{e^{s_{i,i}} + \sum_{s \;\in\; \text{top}k( \mathcal{N}_{i, B_t})} e^{s}} \right)
\label{eq:SNM}
\end{equation}

Fig.~\ref{fig:sampling-methods}~(B) top right shows this scenario. The set of negatives now only comprise the hardest comparisons for the given query. In the diagram, the red shaded negative scores are $\text{top}\,k(\mathcal{N}_{i,B_t})$.

\textbf{Cross-Example Softmax.}
From Equations~\ref{eq:SampledSoftmax} and \ref{eq:SNM} and the illustrations in Fig.~\ref{fig:sampling-methods}~(B) it becomes clear that Sampled Softmax captures the distance of documents only relative with respect to a given query. Thus, distances in the learned vector space are not comparable across queries. 
To encourage global calibration such that distance can be used as an absolute measure of relevance, we propose \emph{Cross-Example Softmax} which extends Softmax by introducing cross-example negatives. The proposed loss
encourages that all queries are closer to their matching documents than all queries are to all irrelevant documents. 
Specifically, let $\mathcal{N}_{B_t}$ be the pairwise comparisons between all queries in batch $B_t$ and the documents of the same batch which they are not related to. Since queries are assumed to be only related to their respective document, this corresponds to all off-diagonal entries in $S$. Formally, 
\begin{equation}
    \mathcal{N}_{B_t} = \bigcup_{i \in [1,...,N]} \mathcal{N}_{i,B_t}
\end{equation}

With this we can formally define Cross-Example Softmax Cross-Entropy as
\begin{equation}
    \mathcal{L}_{B_t} = - \frac{1}{N} \sum_{i=1}^{N}
\log\left(\frac{e^{s_{i,i}}}{e^{s_{i,i}} + \sum_{s \;\in\; \mathcal{N}_{B_t}} e^{s}} \right)
\end{equation}

\begin{figure*}[t]
\includegraphics[width=\linewidth]{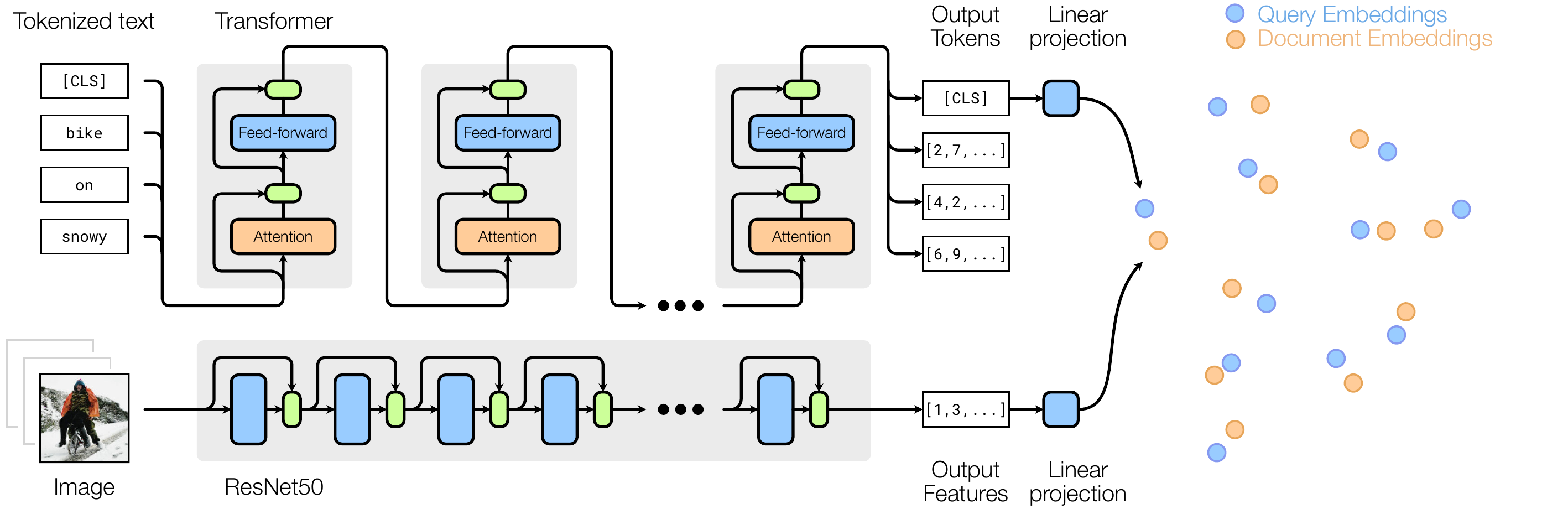}
\caption{\textbf{The Text-to-Image model} comprises two encoders. Text captions are encoded using a 12-layer Transformer. The output of the CLS token is projected into the embedding space and L2 normalized. Images are encoded using a 50-layer Residual Network followed by a linear projection and L2 normalization. The relevance is then measured via dot product in the embedding space.}
\label{fig:overview}
\vspace{-5pt}
\end{figure*}

Fig.~\ref{fig:sampling-methods}~(B) bottom left illustrates that for Cross-Example Softmax the loss for a single query includes all negative scores from $\mathcal{N}_{B_t}$, even query/document pairs from different queries.

\textbf{Cross-Example Negative Mining.}
We can now extend Stochastic Negative Mining to mine for the hardest negative comparisons across the entire batch. Akin to the formulation above, let $\text{top}k( \mathcal{N}_{i, B_t})$ be the set of the top $k$ largest scores within the set of negative scores of the entire batch. With this we can define the Cross-Example Negative Mining loss as 
\begin{equation}
    \mathcal{L}_{B_t} = - \frac{1}{N} \sum_{i=1}^{N}
\log\left(\frac{e^{s_{i,i}}}{e^{s_{i,i}} + \sum_{s \;\in\; \text{top}k(\mathcal{N}_{B_t})} e^{s}} \right)
\end{equation}

This loss is illustrated in Fig.~\ref{fig:sampling-methods}~(B) bottom right. 
Note that negative scores for each query are mined from the entire batch. This means that the set of mined scores could contain all negative scores from some queries, like row~1 in the figure, and no negative scores from others, like row~3 in the figure. 

\section{Experiments}
We perform a series of experiments to evaluate the text-to-image retrieval performance of the proposed Cross-Example Softmax. We use two datasets, i.e., Conceptual Captions~\cite{conceptualcaptions} and Flickr30k~\cite{flickr30k}.
As baselines, we compare to sampled softmax~\cite{sampledsoftmax} and sampled softmax with in-batch negative mining~\cite{snm}, as well as two angular margin-based similarity learning baselines~\cite{lsoftmax,angular-softmax}.

\subsection{Text-To-Image Model}
For our text-to-image retrieval application we learn two separate encoders, limiting the interaction between text and images to a dot product.
This allows to efficiently score a query against a large pre-computed database of image representations using approximate nearest neighbors (ANN), e.g.,~\cite{ChenW18,faiss,wu2017multiscale}. This is in contrast to \emph{cross-attention} such as ViLBERT~\cite{vilbert}, where the interaction between text and image relies on heavy computation and thus does not scale to the retrieval setting.



\textbf{Text Encoder.}
As text encoder $f_{text}$ we use a 12-layer Transformer~\cite{transformer}. Specifically, we use the publicly available pre-trained \verb+BERT_Base+~\cite{bert}. The inputs to the model are word piece tokenized image captions padded to a maximum sequence length of 32. As final vector representations for the caption, we use the 768-dimensional vector representing the \verb+[CLS]+ token and add a dimensionality reduction to 128 dimensions without any bias or activation.

\textbf{Image Encoder.}
As image encoder $f_{image}$ we use a 50-layer ResNet~\cite{resnet} pretrained on the classic ILSVRC2012 classification task. As final vector representations for the image, we replace the last layer with a dimensionality reduction to 128 dimensions without any bias or activation.

\textbf{Relevance Score}. 
Representations are compared using cosine similarity. Since cosine similarities lie within the range from -1 to 1, the resulting dynamic range is too small when used with softmax. We follow~\cite{heatedupsoftmax} and add a temperature $\lambda$, such that the final comparison is  $s_{i,j} = \langle \lambda \frac{\mathbf{x}_i}{|\mathbf{x}_i|}, \lambda \frac{\mathbf{y}_j}{|\mathbf{y}_j|} \rangle$.

\subsection{Training Details}
We use the Conceptual Captions~\cite{conceptualcaptions} dataset to train our text-to-image model. The dataset is set of images from the web along with \verb+alt+ text descriptions. 
The training split includes 3,318,333 and the test split 12,559 image/text pairs. 
\begin{table*}[t]
\small
	\centering
	\tiny

		\vspace{-5pt}
		\begin{tabular}{@{}lc|ccc|cccc@{}} \toprule
		    & & \multicolumn{3}{c|}{Recall on Test Set} & \multicolumn{4}{c}{+3.3M distractors} \\
			\textbf{Model} & \textbf{PR AUC} & \textbf{@1} & \textbf{@5} & \textbf{@10} & \textbf{@1} & \textbf{@5} & \textbf{@10}& \textbf{@100}\\ \midrule
			Sampled Softmax~\cite{sampledsoftmax}  & 14.61 & 25.87&	50.67&	61.12 & 1.38 &	3.99 &	6.16 & 20.44\\ 
			L-Softmax~\cite{lsoftmax}  & 14.03 & 26.18 &	50.78 & 	61.25 & 1.38 &	4.20 &	6.33 & 20.72\\ 
			A-Softmax~\cite{sphereface}  & 14.11 & 26.20 &	\textbf{50.98} & 	\textbf{61.48} & 1.38 &	4.18 &	6.38 & 20.82\\ 
			SNM~\cite{snm}  & 14.80 & 26.08 &	50.71& 	{61.41} & 1.28 &	3.97 &	6.08 & 20.63\\ 
			\midrule
			\textbf{CE-Softmax}  & \textbf{20.12} & \textbf{26.95} &	50.65 &	61.18 & 1.55 &	4.51 &	6.83 & 21.54\\
			\textbf{CE Negative Mining}  & 20.09 & 26.91 &	{50.85} &	61.21 & \textbf{1.57} &	\textbf{4.58} &	\textbf{6.94} & \textbf{21.62}\\
			\bottomrule
		\end{tabular}
		\caption{\label{tab:conceptual}
		\textbf{Retrieval Results on Conceptual Captions.} Left: Cross-Example Softmax improves score calibration as shown by the increase in PR-AUC. Middle: When retrieving images from the test set only, Cross-Example Softmax outperforms Sampled Softmax and the margin-based baselines for Recall@1. 
		Right: With a large number of distractor images, Cross-Example Softmax clearly outperforms vanilla Sampled Softmax across all levels of $k$. Cross-Example Negative Mining further improves upon these results. All numbers shown are the average of 5 runs.}
		\vspace{-5pt}
\end{table*}
Our models are trained with batch size 512 for 25,000 steps.
For models with negative mining, we select the 50\% largest scores from the respective negative sets.
Due to the different architectures in the two encoders, we use two separate optimizers~\cite{acclip}. The Transformer uses AdamW with a learning rate of 1e-4 that is linearly decayed to 0 after a warmup period of 1,500 steps. The ResNet uses SGD with momentum of 0.9 and a linear learning rate decay schedule starting at 3e-3 and ending at 5e-4. The output logits are scaled with a factor of $\lambda=20$.

\subsection{Retrieval Results on Conceptual Captions}
We now evaluate retrieval performance and score calibration of the learned embeddings.

\textbf{Recall@k.} We evaluate the retrieval performance of the different models
%
on two retrieval sets of varying size.
First, we focus solely on the test set, i.e., for all test captions we retrieve the most relevant image from the set of all test images.
Second, for a more realistic and challenging setting we add the 3.3M training images as distractors to the retrieval set.
%
Table~\ref{tab:conceptual} shows Recall@k for the first setting in the middle column and for the second setting on the right side.
%
From the results, we observe that for the first setting, Cross-Example Softmax outperforms vanilla Sampled Softmax and the margin-based baselines for Recall@1. For recall at larger $k$ we do not see large differences between the models.
For the second setting, we observe that Cross-Example Softmax clearly outperforms all baselines across all levels of $k$. Cross-Example Negative Mining further improves upon these results and consistently achieves the best performance across all models.


\textbf{Precision-Recall and AUC.}
To measure global score calibration, we now evaluate the Precision-Recall curve, measuring the precision of all returned results at a specified global recall threshold.
To generate PR curves, we compute the pairwise similarities between all captions and images in the test set and sort them by their similarity score.
%
\begin{figure}[t]
\centering
\begin{subfigure}{0.45\linewidth}
\centering
\includegraphics[width=1\linewidth]{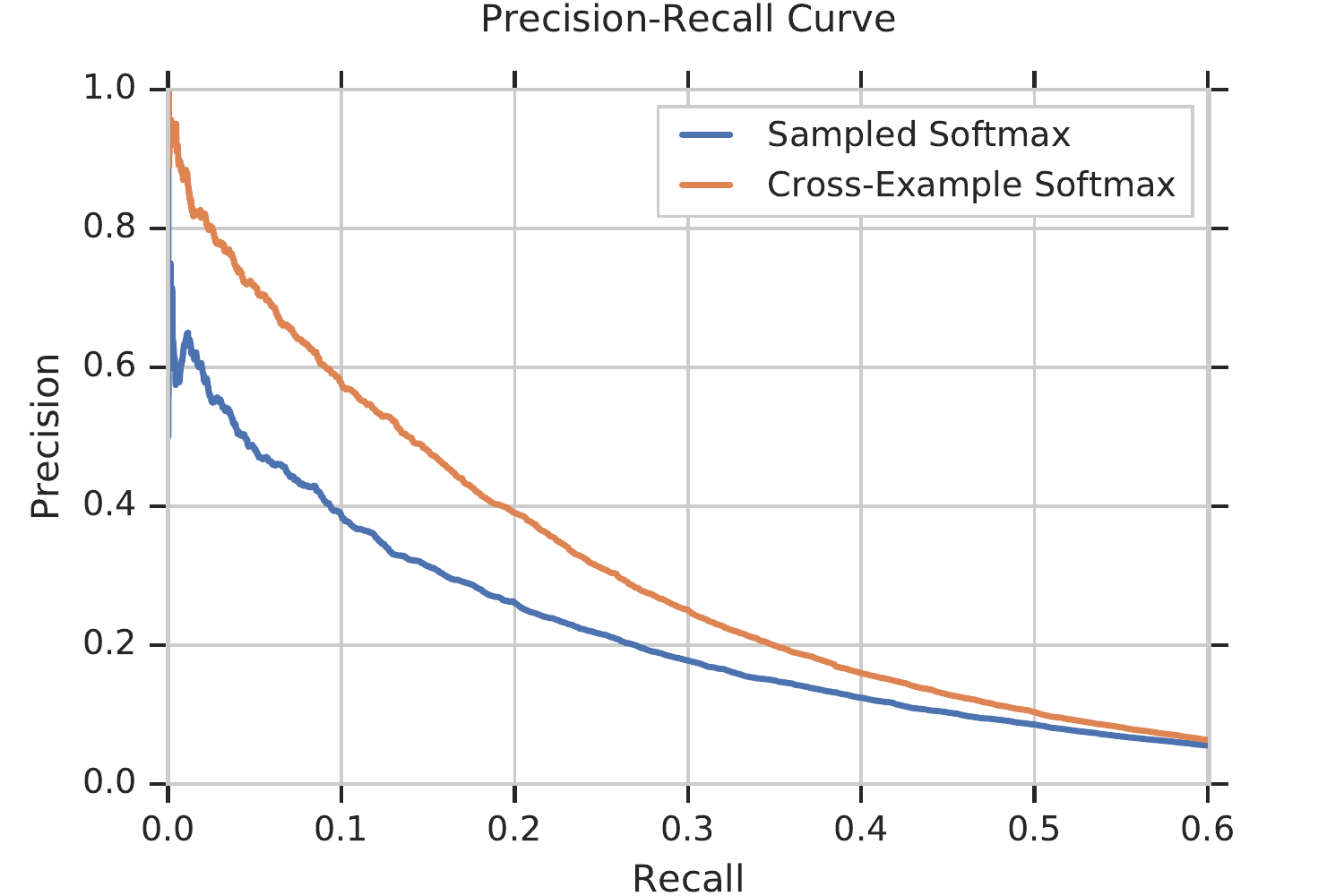}
\end{subfigure}
\begin{subfigure}{0.4\linewidth}
\centering
\includegraphics[width=1\linewidth]{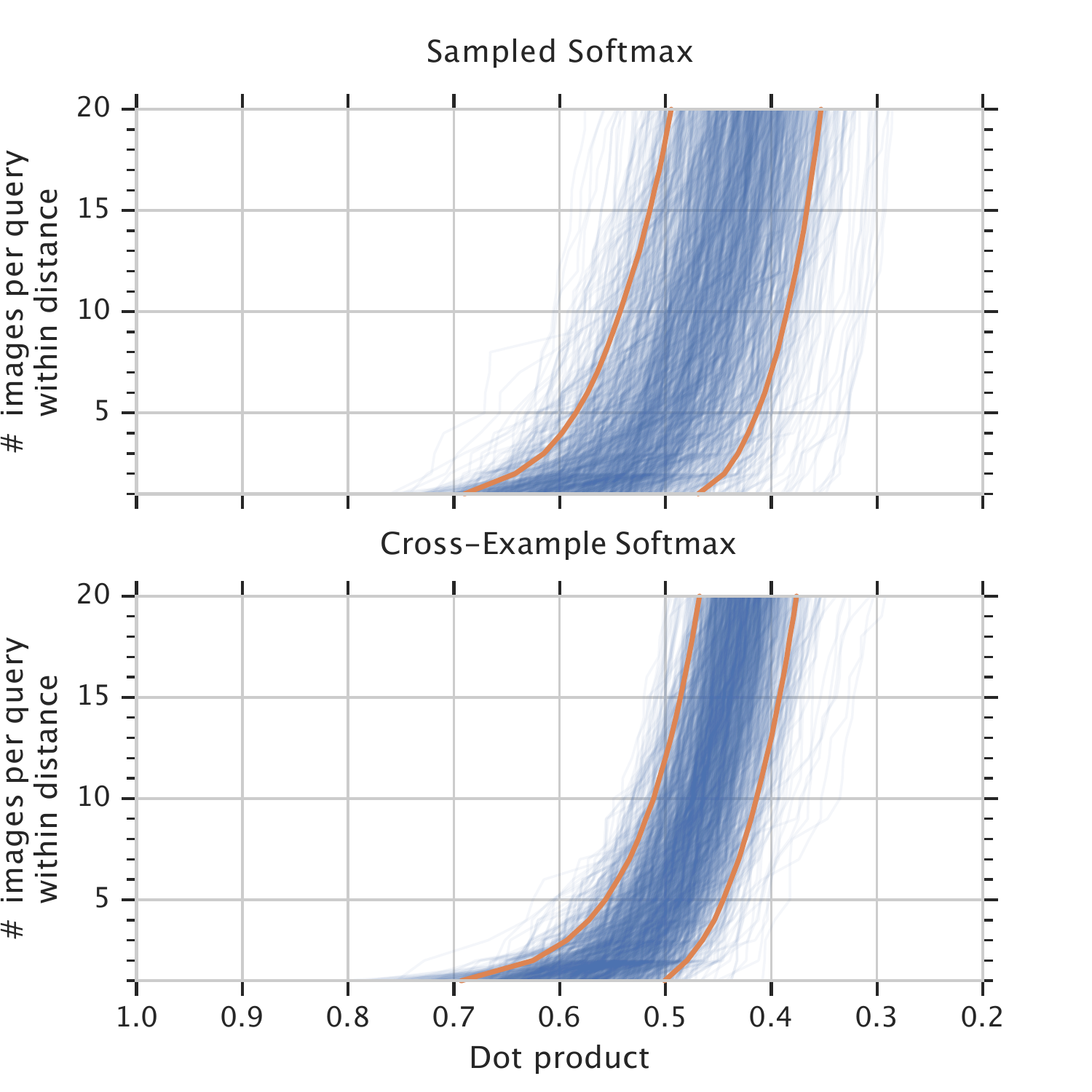}
\end{subfigure}
\vspace{-10pt}
\caption{
\label{fig:pr-curve}
\label{fig:hair}
\textbf{Left: Precision-Recall curves} on the Conceptual Captions test set images for Sampled Softmax and Cross-Example Softmax. Incorporating negative comparisons across examples leads to much better global calibration and correspondingly higher AUC.
\textbf{Right: Comparison of score distributions.}
Each blue line is one of 1,000 randomly sampled queries from the test set, showing the number of neighbors ($y$-axis) at a given similarity to the query ($x$-axis).
The orange lines indicate the 5th and 95th percentile, i.e., 90\% of queries fall in between. The result demonstrates how Cross-Example Softmax leads to a more globally-calibrated similarity metric. 
}

\end{figure}
The Precision-Recall curves for Sampled Softmax and Cross-Example Softmax are shown in Figure~\ref{fig:pr-curve}~(left). Further, the PR AUC for all methods is shown on the left of Table~\ref{tab:conceptual}.
The results clearly indicate that Cross-Example Softmax improves the global score calibration with an increase in PR-AUC from 0.14 to 0.21.
The PR curves in Figure~\ref{fig:pr-curve} highlight that much of the improvement comes from a better separation between the positives and the hardest negative documents. 
This highlights the effectiveness of directly optimizing for a better separation between positives and the entire distribution of all negative scores.

\subsection{Zero-Shot Retrieval Results on Flickr30k}

We now perform a set of zero-shot experiments on Flickr30k to evaluate the generalization of the learned embeddings. We use the zero-shot setting and test split from \cite{vilbert}, which contains 1,000 images, each with 5 captions. Following the procedure of previous work, we compare all 5,000 captions to all 1,000 images in the Flickr30k test set in order to retrieve the most relevant images to each query. Note that we do not perform any dataset-specific fine-tuning on Flickr30k; all of our results are shown in the zero-shot setting of training on conceptual captions and testing on Flickr30k.

\begin{table}[t]
\small
	\centering
	\tiny
		\vspace{-3pt}
		\begin{tabular}{@{}lccc@{}} \toprule
			\textbf{Model} & \textbf{Recall@1} & \textbf{Recall@5} & \textbf{Recall@10}\\ \midrule
			ViLBERT~\cite{vilbert} & 31.86&	61.12&	72.80\\ \midrule
			Sampled Softmax~\cite{sampledsoftmax}  & 29.22 &	55.70 &	67.20 \\ 
			L-Softmax~\cite{lsoftmax}  & 29.12 &	55.79 &	66.95 \\
			A-Softmax~\cite{sphereface}  & 28.74 &	55.80 &	66.96 \\
			SNM~\cite{snm}  & 29.07 &	55.95 &	67.27\\
			\midrule
			\textbf{CE-Softmax}  & 29.94 & 	56.83 &	67.78\\
			\textbf{CE Negative Mining}  & \textbf{30.49} &	\textbf{57.04} &	\textbf{68.06}\\
			\bottomrule
		\end{tabular}
		\vspace{2pt}
		\caption{\label{tab:flickr}
		\textbf{Retrieval Results on Flickr30k.} Cross-Example softmax clearly outperforms Sampled Softmax across all recall levels. Moreover, Cross-Example Negative Mining improves further upon the results. Numbers shown are the average accross 5 runs.}
\end{table}

Table~\ref{tab:flickr} shows the retrieval performance.
Besides the methods presented above, the table further includes a comparison to ViLBERT~\cite{vilbert}. However, as mentioned earlier, note that ViLBERT is a cross-attention model and thus does not scale to large scale retrieval.
Similarly to the Conceptual Captions 3M results above, we find that 
Cross-Example Softmax outperforms the baselines across all levels of $k$.
Focusing on the hardest examples with Cross-Example Negative Mining further improves recall, demonstrating the effectiveness of the proposed methods.
It is interesting that our model performs roughly on-par with cross-attention models like ViLBERT that rely on combining visual and text features early in the model, ruling them infeasible for retrieval tasks over large-scale databases.


\begin{figure*}[t]
\centering
\includegraphics[width=0.7\linewidth]{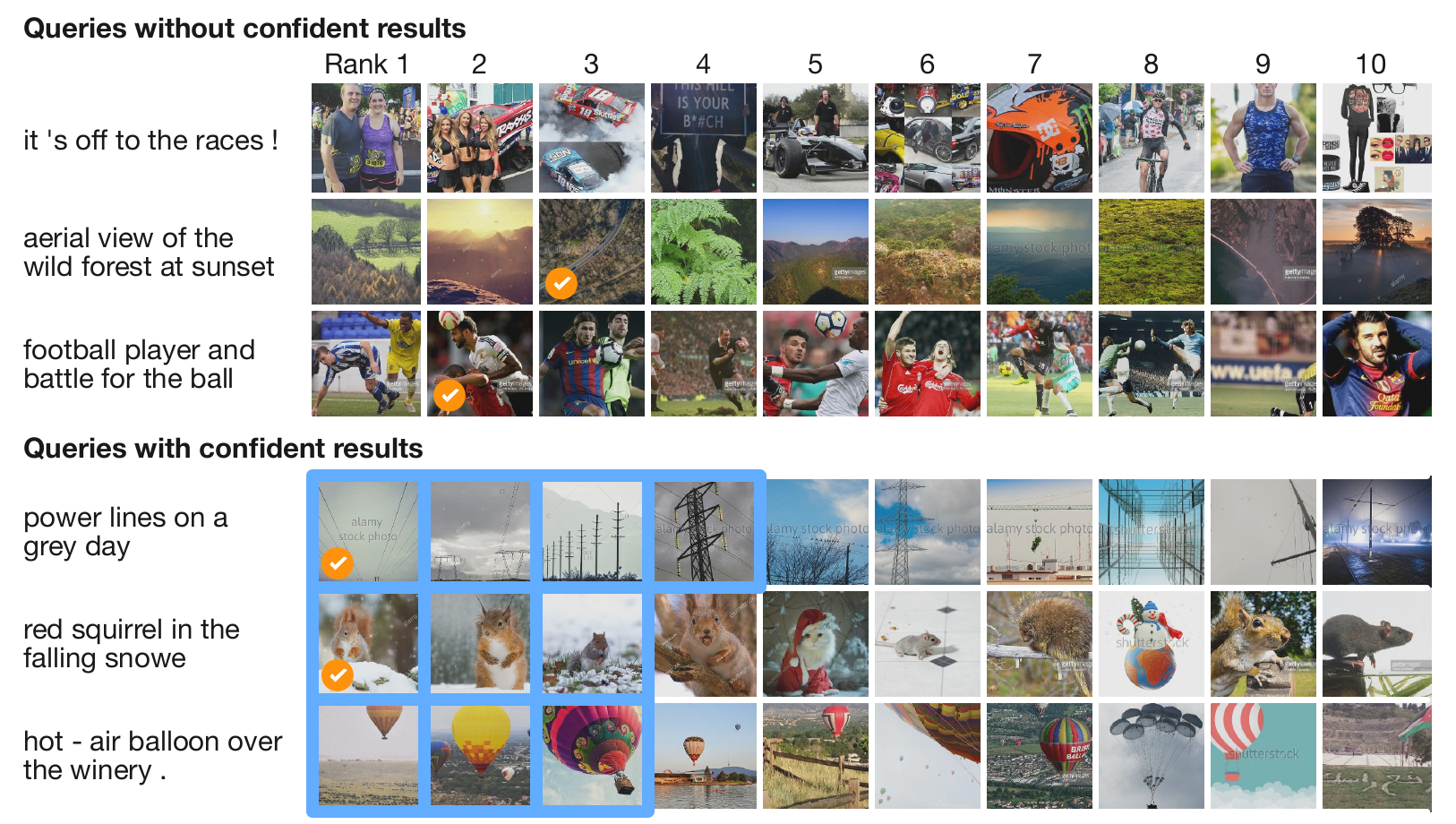}
\caption{\label{fig:examples}
\textbf{Example queries and top retrieval results} from Cross-Example Softmax on the Conceptual Captions test set. Retrieved ground truth image have an orange check mark. Images with high relevance score, i.e., $\geq 0.6$, have a blue outline. Top: queries without any high relevance score results. Bottom: queries that contain results with high relevance scores. Queries without high confidence result tend to be vague or not visually descriptive. Queries with high scoring results tend to be more descriptive and images with high scores are mostly plausibly correct. The last example shows a failure case, where the model ignores the `winery' token and solely focuses on the hot air balloon.
}
\label{fig:examples}
\vspace{-5pt}
\end{figure*}

\subsection{Qualitative Analysis}

\textbf{Score distributions.}
First we look at the effect that Cross-Example Softmax has on the score distribution of retrieved documents. 
To analyze the distributions, Fig.~\ref{fig:hair}~(right) shows the number of neighbors ($y$-axis) at a given similarity to the query ($x$-axis) for 1,000 randomly sampled queries from the Conceptual Captions test set. Each blue line represents one query. 
Blue lines on the leftmost side represent queries whose top retrieved results have higher similarity scores, and the lines on the rightmost side represent queries whose top results have comparatively low similarity. 
The orange lines indicate the 5th and 95th percentile, i.e., 90\% of queries fall in between. The result demonstrates how Cross-Example Softmax leads to a more globally-calibrated similarity metric. 

\textbf{Example results.}
Fig.~\ref{fig:examples} shows example queries and their top retrieved results. Specifically, it shows three queries where all returned results have low confidence scores and three queries that return many high confidence results. Those queries would show on the rightmost and leftmost sides within the plot of Fig.~\ref{fig:hair} respectively. Generally, we observe that queries without any highly confident results are often vague or not visually descriptive.
On the other hand, queries with high scoring results tend to be more descriptive and images with high scores are mostly plausible results. 

\section{Conclusion}
In this work, we proposed Cross-Example Softmax, wherein given a batch of inputs and labels, the loss encourages that the score of a correct input/label pair is higher than all incorrect input/label pairs, even across multiple inputs. By extending the loss to all input/label pairs, we ensure that the relevance score is more globally calibrated and thus becomes more interpretable as an absolute measure of relevance. We further introduce Cross-Example Negative Mining, where the loss is focused on the hardest incorrect input/label pairs. Empirically, we showed that the proposed methods effectively improve global calibration as well as retrieval performance.
This work opens up numerous paths for future work. From a practitioner’s point of view, it might be exciting to extend this work towards multiclass classification and object detection. From a robustness perspective, it would be intriguing to study the possible impact of the proposed method on susceptibility towards adversarial examples.

\clearpage
\clearpage
\section*{Broader Impact}

Interpretability is a key consideration of machine learning applications. In this work, we propose a novel method for making distances in embedding-based retrieval methods more interpretable. We emphasize that this only addresses one specific aspect of interpretability and only for retrieval. 

Further, the general ethical concerns caused by improvements in image recognition do also apply to this work. Although the methods proposed in this paper are motivated to improve model interpretability, they could lead unforeseen applications.

\bibliographystyle{plainnat}
\bibliography{references}

\end{document}